\documentclass[10pt,twocolumn,letterpaper]{article}

\usepackage{iccv}
\usepackage{times}
\usepackage{amsmath}
\usepackage{amssymb}
\usepackage{booktabs}
\usepackage{graphicx}
\usepackage{soul}
\usepackage{subfig}
\usepackage{tabularx}

\makeatletter
\@namedef{ver@everyshi.sty}{}
\makeatother
\usepackage{tikz}
\usepackage{pgf, pgfsys, pgffor}
\usepackage{pgfplots}

\usepackage[pagebackref=true, breaklinks=true, letterpaper=true, colorlinks, bookmarks=false]{hyperref}

\def\iccvPaperID{1148} 
\iccvfinalcopy 

\newcolumntype{Y}{>{\centering\arraybackslash}X}

\ificcvfinal\fi
\begin{document}

\title{Pix2Vox: Context-aware 3D Reconstruction from Single and Multi-view Images}

\author{Haozhe Xie$^\dag$\hspace{0.05in}
Hongxun Yao$^\dag$\hspace{0.05in}
Xiaoshuai Sun$^\dag$\hspace{0.05in}
Shangchen Zhou$^\ddag$\hspace{0.05in}
Shengping Zhang$^\dag$$^\S$\\
$^\dag$Harbin Institute of Technology\hspace{0.1in}
$^\ddag$SenseTime Research 
$^\S$Peng Cheng Laboratory \\
$^\dag${\tt\small\{hzxie, h.yao, xiaoshuaisun, s.zhang\}@hit.edu.cn} 
$^\ddag${\tt\small zhoushangchen@sensetime.com} \\
{\small\url{https://haozhexie.com/project/pix2vox}}
}

\maketitle

\begin{abstract}
Recovering the 3D representation of an object from single-view or multi-view RGB images by deep neural networks has attracted increasing attention in the past few years.
Several mainstream works (e.g., 3D-R2N2) use recurrent neural networks (RNNs) to fuse multiple feature maps extracted from input images sequentially.
However, when given the same set of input images with different orders, RNN-based approaches are unable to produce consistent reconstruction results.
Moreover, due to long-term memory loss, RNNs cannot fully exploit input images to refine reconstruction results.
To solve these problems, we propose a novel framework for single-view and multi-view 3D reconstruction, named Pix2Vox.
By using a well-designed encoder-decoder, it generates a coarse 3D volume from each input image.
Then, a context-aware fusion module is introduced to adaptively select high-quality reconstructions for each part (e.g., table legs) from different coarse 3D volumes to obtain a fused 3D volume.
Finally, a refiner further refines the fused 3D volume to generate the final output.
Experimental results on the ShapeNet and Pix3D benchmarks indicate that the proposed Pix2Vox outperforms state-of-the-arts by a large margin.
Furthermore, the proposed method is 24 times faster than 3D-R2N2 in terms of backward inference time.
The experiments on ShapeNet unseen 3D categories have shown the superior generalization abilities of our method.
\end{abstract}

\vspace{-4 mm}
\section{Introduction}

\begin{figure}[!t]
  \centering
  \resizebox{\linewidth}{!} {
    \begin{tikzpicture}
      \definecolor{red}{rgb}{0.812, 0, 0.118}
      \definecolor{yellow}{rgb}{0.980, 0.761, 0.306}
      \definecolor{orange}{rgb}{0.965, 0.467, 0.192}
      \definecolor{green}{rgb}{0.086, 0.365, 0.133}
      \definecolor{lightgreen}{rgb}{0.412, 0.741, 0.392}
      \definecolor{pink}{rgb}{0.933, 0.271, 0.565}

      \begin{axis}[
        xlabel = {Forward Inference Time (ms)},
        ylabel = {Intersection over Union (IoU)},
        ylabel style = {at = {(0.025, 0.5)}},
        xtick = {0, 10, 20, 30, 40, 50, 60, 70, 80, 90},
        ytick = {0.505, 0.54, 0.56, 0.58, 0.60, 0.62, 0.64, 0.66},
        yticklabels = {, 0.54, 0.56, 0.58, 0.60, 0.62, 0.64, 0.66},
        xmajorgrids = true,
        ymajorgrids = true,
        grid style = dashed,
        enlargelimits = 0.1,
        scatter/classes = {
          method={mark=*, white},
          ref={mark=*, transparent}
        }
      ]
      \addplot [
        scatter,
        only marks,
        scatter src=explicit symbolic,
      ] coordinates {
        (10, 0.665)     [ref]    
        (90, 0.505)     [ref]    

        (9.25, 0.634)   [method] 
        (9.90, 0.661)   [method] 
        (73.35, 0.560)  [method] 
        (85.73, 0.640)  [method] 
        (37.90, 0.596)  [method] 
      };

      \draw[white, ultra thick, fill=lightgreen, opacity=0.8](axis cs: 9.90, 0.661)  circle (0.832 cm);
      \node [right] at (axis cs:  20, 0.661) {\textbf{Pix2Vox-A}};
      
      \draw[white, ultra thick, fill=green, opacity=0.8](axis cs: 9.25, 0.634)  circle (0.365 cm);
      \node [below] at (axis cs:  12.5, 0.624) {\textbf{Pix2Vox-F}};

      \draw[white, ultra thick, fill=orange, opacity=0.8](axis cs: 85.73, 0.640)  circle (1.001 cm);
      \node [below]  at (axis cs:  85.73, 0.610) {PSGN};

      \draw[white, ultra thick, fill=pink, opacity=0.8](axis cs: 73.35, 0.560)  circle (0.588 cm);
      \node [left]  at (axis cs:  66, 0.560) {3D-R2N2};

      \draw[white, ultra thick, fill=yellow, opacity=0.8](axis cs: 37.90, 0.596)  circle (0.427 cm);
      \node [right] at (axis cs:  43, 0.596) {OGN};
      
      \draw[white, ultra thick, fill=lightgray, opacity=0.6](axis cs: 57, 0.505)  circle (0.325 cm);
      \node[above] at (axis cs:  57, 0.525)  {5 M};
      \draw[white, ultra thick, fill=lightgray, opacity=0.6](axis cs: 46, 0.505)  circle (0.4 cm);
      \node[above] at (axis cs:  46, 0.525)  {10 M};
      \draw[white, ultra thick, fill=lightgray, opacity=0.6](axis cs: 30.5, 0.505)  circle (0.649 cm);
      \node[above] at (axis cs:  30.5, 0.525) {50 M};
      \draw[white, ultra thick, fill=lightgray, opacity=0.6](axis cs: 10, 0.505) circle (0.8 cm);
      \node[above] at (axis cs:  10, 0.525) {100 M};
      \node[right] at (axis cs:  62.5, 0.505)  {\# Parameters};
      \end{axis}
    \end{tikzpicture}
  }
  \caption{Forward inference time, model size, and IoU of state-of-the-arts and our methods for single-view 3D reconstruction on the ShapeNet testing set. The radius of each circle represents the size of the corresponding model. Pix2Vox outperforms state-of-the-arts in forward inference time and reaches the best balance between accuracy and model size.}
  \label{fig:performance-comparison}
  \vspace{-3 mm}
\end{figure}
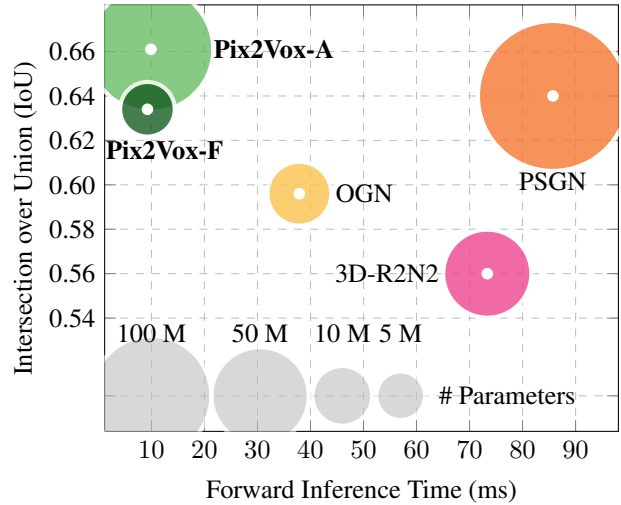

\begin{figure*}
  \centering
  \resizebox{\linewidth}{!} {
    \includegraphics{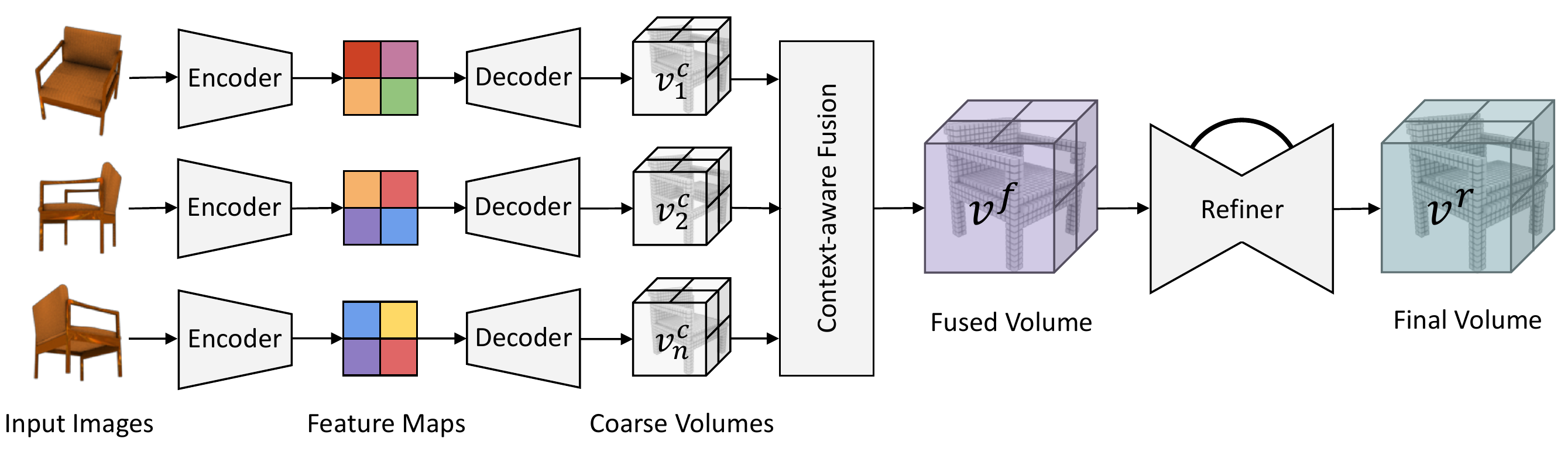}
  }
  \caption{An overview of the proposed Pix2Vox. The network recovers the shape of 3D objects from arbitrary (uncalibrated) single or multiple images. The reconstruction results can be refined when more input images are available. Note that the weights of the encoder and decoder are shared among all views.}
  \label{fig:overview}
  \vspace{-3 mm}
\end{figure*}

3D reconstruction is an important problem in robotics, CAD, virtual reality and augmented reality.
Traditional methods, such as Structure from Motion (SfM) \cite{DBLP:journals/acta/OnurVRA17} and Simultaneous Localization and Mapping (SLAM) \cite{DBLP:journals/air/Fuentes-PachecoAR15},  match image features across views.
However, establishing feature correspondences becomes extremely difficult when multiple viewpoints are separated by a large margin due to local appearance changes or self-occlusions \cite{DBLP:journals/ijcv/Lowe04}.
To overcome these limitations, several deep learning based approaches, including 3D-R2N2 \cite{DBLP:conf/eccv/ChoyXGCS16}, LSM \cite{DBLP:conf/nips/KarHM17}, and 3DensiNet \cite{DBLP:conf/mm/WangWF17}, have been proposed to recover the 3D shape of an object and obtained promising results.

To generate 3D volumes, 3D-R2N2 \cite{DBLP:conf/eccv/ChoyXGCS16} and LSM \cite{DBLP:conf/nips/KarHM17} formulate multi-view 3D reconstruction as a sequence learning problem and use recurrent neural networks (RNNs) to fuse multiple feature maps extracted by a shared encoder from input images.
The feature maps are incrementally refined when more views of an object are available.
However, RNN-based methods suffer from three limitations.
First, when given the same set of images with different orders, RNNs are unable to estimate the 3D shape of an object consistently results due to permutation variance \cite{DBLP:conf/iclr/VinyalsBK16}.
Second, due to long-term memory loss of RNNs, the input images cannot be fully exploited to refine reconstruction results \cite{DBLP:conf/icml/PascanuMB13}.
Last but not least, RNN-based methods are time-consuming since input images are processed sequentially without parallelization \cite{DBLP:conf/icassp/HwangS15}.
 
To address the issues mentioned above, we propose Pix2Vox, a novel framework for single-view and multi-view 3D reconstruction that contains four modules: encoder, decoder, context-aware fusion, and refiner.
The encoder and decoder generate coarse 3D volumes from multiple input images in parallel, which eliminates the effect of the orders of input images and accelerates the computation.
Then, the context-aware fusion module selects high-quality reconstructions from all coarse 3D volumes and generates a fused 3D volume, which fully exploits information of all input images without long-term memory loss.
Finally, the refiner further correct wrongly recovered parts of the fused 3D volumes to obtain a refined reconstruction.
To achieve a good balance between accuracy and model size, we implement two versions of the proposed framework: Pix2Vox-F and Pix2Vox-A (Figure \ref{fig:performance-comparison}).

The contributions can be summarized as follows:

\begin{itemize}
  \vspace{-1.5 mm}
  \item We present a unified framework for both single-view and multi-view 3D reconstruction, namely Pix2Vox. We equip Pix2Vox with well-designed encoder, decoder, and refiner, which shows a powerful ability to handle 3D reconstruction in both synthetic and real-world images.
  \vspace{-1.5 mm}
  \item We propose a context-aware fusion module to adaptively select high-quality reconstructions for each part from different coarse 3D volumes in parallel to produce a fused reconstruction of the whole object. To the best of our knowledge, it is the first time to exploit context across multiple views for 3D reconstruction.
  \vspace{-1.5 mm}
  \item Experimental results on the ShapeNet \cite{DBLP:conf/cvpr/WuSKYZTX15} and Pix3D \cite{DBLP:conf/cvpr/Sun0ZZZXTF18} datasets demonstrate that the proposed approaches outperform state-of-the-art methods in terms of both accuracy and efficiency. Additional experiments also show its strong generalization abilities in reconstructing unseen 3D objects.
\end{itemize}

\section{Related Work}

\begin{figure*}
  \centering
  \resizebox{\linewidth}{!} {
    \includegraphics{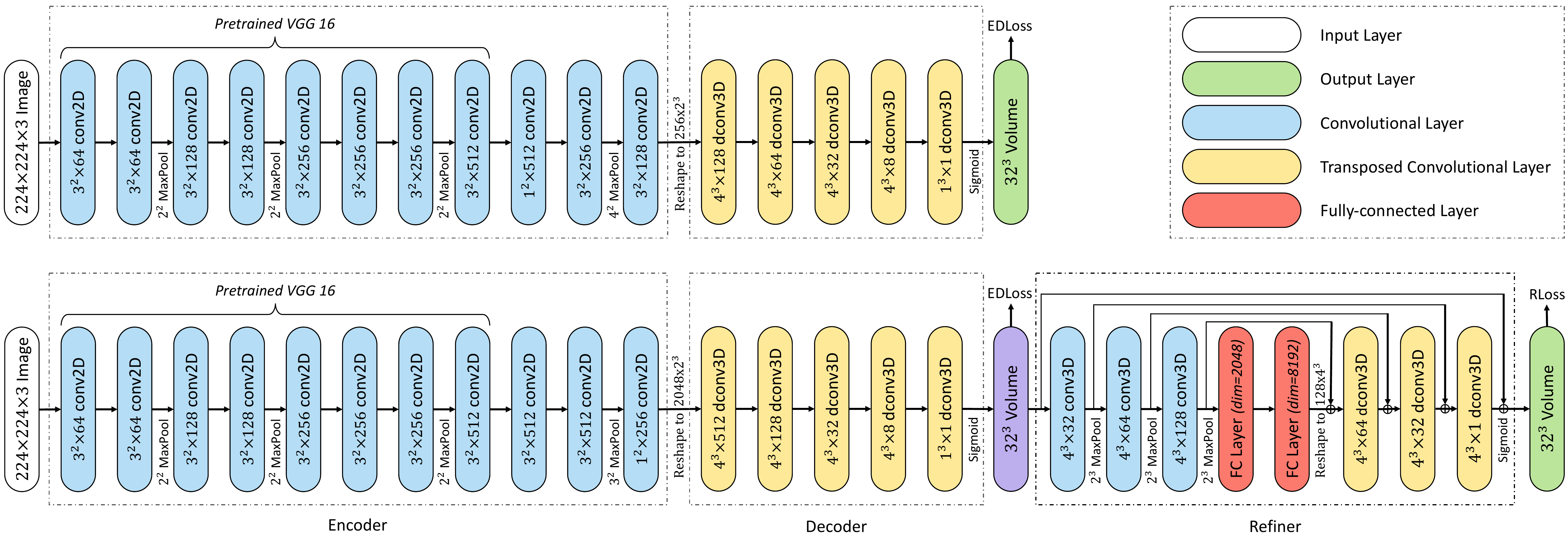}
  }
  \caption{The network architecture of (top) Pix2Vox-F and (bottom) Pix2Vox-A. The EDLoss and the RLoss are defined as Equation \ref{eq:bce-loss}. To reduce the model size, the refiner is removed in Pix2Vox-F.}
  \label{fig:network}
  \vspace{-2 mm}
\end{figure*}

\noindent \textbf{Single-view 3D Reconstruction}
Theoretically, recovering 3D shape from single-view images is an ill-posed problem.
To address this issue, many attempts have been made, such as ShapeFromX \cite{DBLP:journals/pami/BarronM15, DBLP:journals/ijcv/SavareseARBP07}, where X may represent silhouettes \cite{DBLP:conf/cvpr/DibraJOZG17}, shading \cite{DBLP:conf/cvpr/RichterR15}, and texture \cite{DBLP:journals/ai/Witkin81}.
However, these methods are barely applicable to use in the real-world scenarios, because all of them require strong presumptions and abundant expertise in natural images \cite{DBLP:journals/access/ZhangLLPL19}. 
With the success of generative adversarial networks (GANs) \cite{DBLP:conf/nips/GoodfellowPMXWOCB14} and variational autoencoders (VAEs) \cite{DBLP:journals/corr/KingmaW13}, 3D-VAE-GAN \cite{DBLP:conf/nips/0001ZXFT16} adopts GAN and VAE to generate 3D objects by taking a single-view image as input.
However, 3D-VAE-GAN requires class labels for reconstruction.
MarrNet \cite{DBLP:conf/nips/0001WXSFT17} reconstructs 3D objects by estimating depth, surface normals, and silhouettes of 2D images, which is challenging and usually leads to severe distortion \cite{DBLP:phdthesis/ucb/Tulsiani18}.
OGN \cite{DBLP:conf/iccv/TatarchenkoDB17} and O-CNN \cite{DBLP:journals/tog/WangLGST17} use octree to represent higher resolution volumetric 3D objects with a limited memory budget.
However, OGN representations are complex and consume more computational resources due to the complexity of octree representations.
PSGN \cite{DBLP:conf/cvpr/FanSG17} and 3D-LMNet \cite{DBLP:conf/bmvc/MandikalLAR18} generate point clouds from single-view images.
However, the points have a large degree of freedom in the point cloud representation because of the limited connections between points.
Consequently, these methods cannot recover 3D volumes accurately \cite{DBLP:conf/eccv/WangZLF18}.

\noindent \textbf{Multi-view 3D Reconstruction}
SfM \cite{DBLP:journals/acta/OnurVRA17} and SLAM \cite{DBLP:journals/air/Fuentes-PachecoAR15} methods are successful in handling many scenarios.
These methods match features among images and estimate the camera pose for each image.
However, the matching process becomes difficult when multiple viewpoints are separated by a large margin.
Besides, scanning all surfaces of an object before reconstruction is sometimes impossible, which leads to incomplete 3D shapes with occluded or hollowed-out areas \cite{DBLP:journals/pami/YangRMTW18}.
Powered by large-scale datasets of 3D CAD models (e.g., ShapeNet \cite{DBLP:conf/cvpr/WuSKYZTX15}), deep-learning-based methods have been proposed for 3D reconstruction.
Both 3D-R2N2 \cite{DBLP:conf/eccv/ChoyXGCS16} and LSM \cite{DBLP:conf/nips/KarHM17} use RNNs to infer 3D shape from single or multiple input images and achieve impressive results.
However, RNNs are time-consuming and permutation-variant, which produce inconsistent reconstruction results.
3DensiNet \cite{DBLP:conf/mm/WangWF17} uses max pooling to aggregate the features from multiple images.
However, max pooling only extracts maximum values from features, which may ignore other valuable features that are useful for 3D reconstruction.

\section{The Method}

\subsection{Overview}

The proposed Pix2Vox aims to reconstruct the 3D shape of an object from either single or multiple RGB images. 
The 3D shape of an object is represented by a 3D voxel grid, where $0$ is an empty cell and $1$ denotes an occupied cell.
The key components of Pix2Vox are shown in Figure \ref{fig:overview}.
First, the encoder produces feature maps from input images.
Second, the decoder takes each feature map as input and generates a coarse 3D volume correspondingly.
Third, single or multiple 3D volumes are forwarded to the context-aware fusion module, which adaptively selects high-quality reconstructions for each part from coarse 3D volumes to obtain a fused 3D volume.
Finally, the refiner with skip-connections further refines the fused 3D volume to generate the final reconstruction result.

\subsection{Network Architecture}

Figure \ref{fig:network} shows the detailed architectures of Pix2Vox-F and Pix2Vox-A.
The former involves much fewer parameters and lower computational complexity, while the latter has more parameters, which can construct more accurate 3D shapes but has higher computational complexity.

\vspace{-3 mm}
\subsubsection{Encoder}
\vspace{-1 mm}
The encoder is to compute a set of features for the decoder to recover the 3D shape of the object.
The first nine convolutional layers, along with the corresponding batch normalization layers and ReLU activations of a VGG16 \cite{DBLP:conf/iclr/SimonyanZ14a} pretrained on ImageNet \cite{DBLP:conf/cvpr/DengDSLL009}, are used to extract a $512 \times 28 \times 28$ feature tensor from a $224 \times 224 \times 3$ image.
This feature extraction is followed by three sets of 2D convolutional layers, batch normalization layers and ELU layers to embed semantic information into feature vectors.
In Pix2Vox-F, the kernel size of the first convolutional layer is $1^2$ while the kernel sizes of the other two are $3^2$.
The number of output channels of the convolutional layer starts with $512$ and decreases by half for the subsequent layer and ends up with $128$.
In Pix2Vox-A, the kernel sizes of the three convolutional layers are $3^2$, $3^2$, and $1^2$, respectively.
The output channels of the three convolutional layers are $512$, $512$, and $256$, respectively.
After the second convolutional layer, there is a max pooling layer with kernel sizes of $3^2$ and $4^2$ in Pix2Vox-F and Pix2Vox-A, respectively.
The feature vectors produced by Pix2Vox-F and Pix2Vox-A are of sizes $2048$ and $16384$, respectively.
 
\vspace{-3 mm}
\subsubsection{Decoder}
\vspace{-1 mm}
The decoder is responsible for transforming information of 2D feature maps into 3D volumes.
There are five 3D transposed convolutional layers in both Pix2Vox-F and Pix2Vox-A.
Specifically, the first four transposed convolutional layers are of a kernel size of $4^3$, with stride of $2$ and padding of $1$.
There is an additional transposed convolutional layer with a bank of $1^3$ filter.
Each transposed convolutional layer is followed by a batch normalization layer and a ReLU activation except for the last layer followed by a sigmoid function.
In Pix2Vox-F, the numbers of output channels of the transposed convolutional layers are $128$, $64$, $32$, $8$, and $1$, respectively.
In Pix2Vox-A, the numbers of output channels of the five transposed convolutional layers are $512$, $128$, $32$, $8$, and $1$, respectively.
The decoder outputs a $32^3$ voxelized shape in the object's canonical view.

\vspace{-3 mm}
\subsubsection{Context-aware Fusion}
\vspace{-1 mm}

\begin{figure}[!t]
  \centering
  \resizebox{\linewidth}{!} {
    \includegraphics{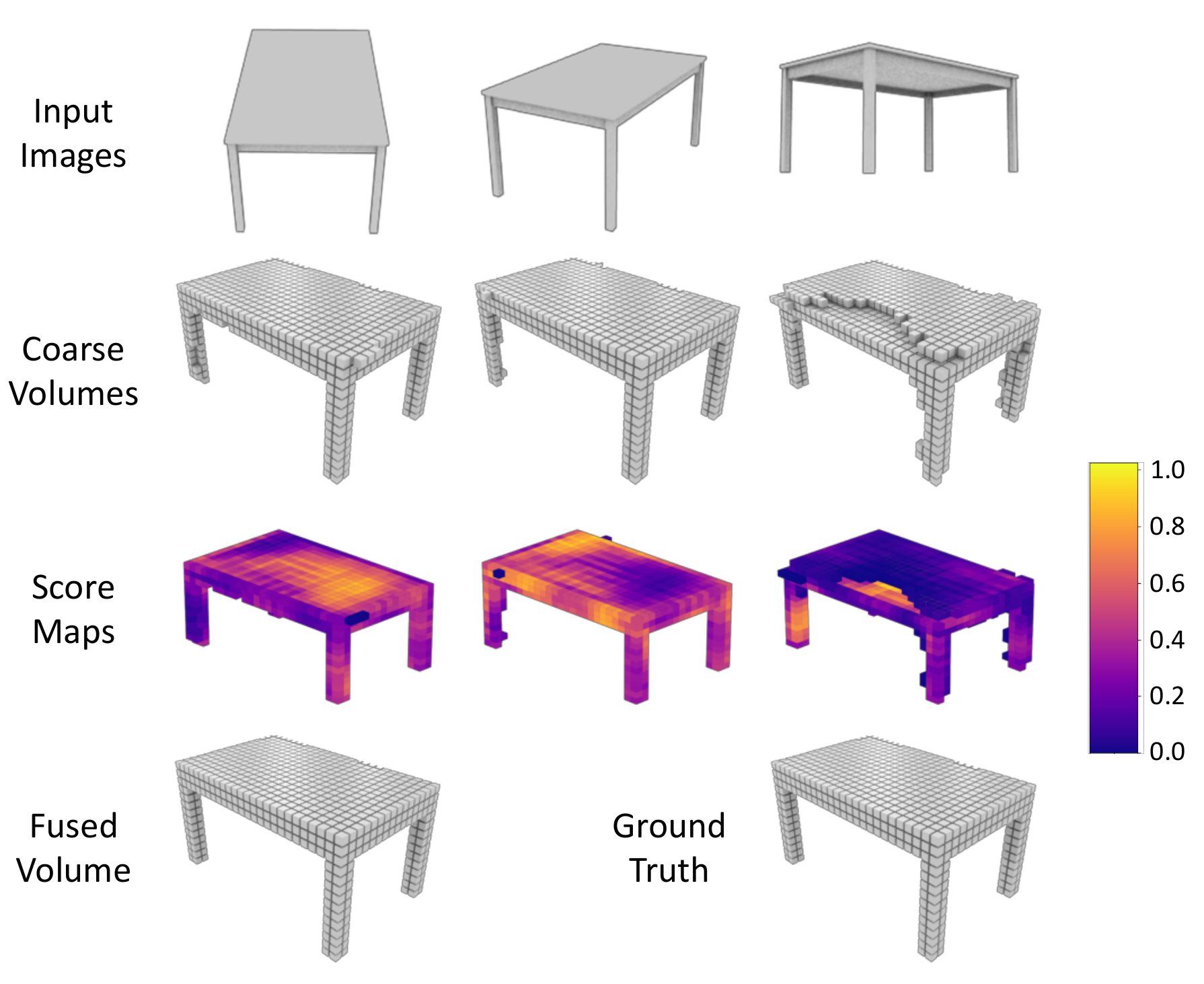}
  }
  \caption{Visualization of the score maps in the context-aware fusion module. The context-aware fusion module generates higher scores for high-quality reconstructions, which can eliminate the effect of the missing or wrongly recovered parts.}
  \label{fig:context-aware-fusion-visualization}
\end{figure}

\begin{figure}
  \centering
  \resizebox{\linewidth}{!} {
    \includegraphics{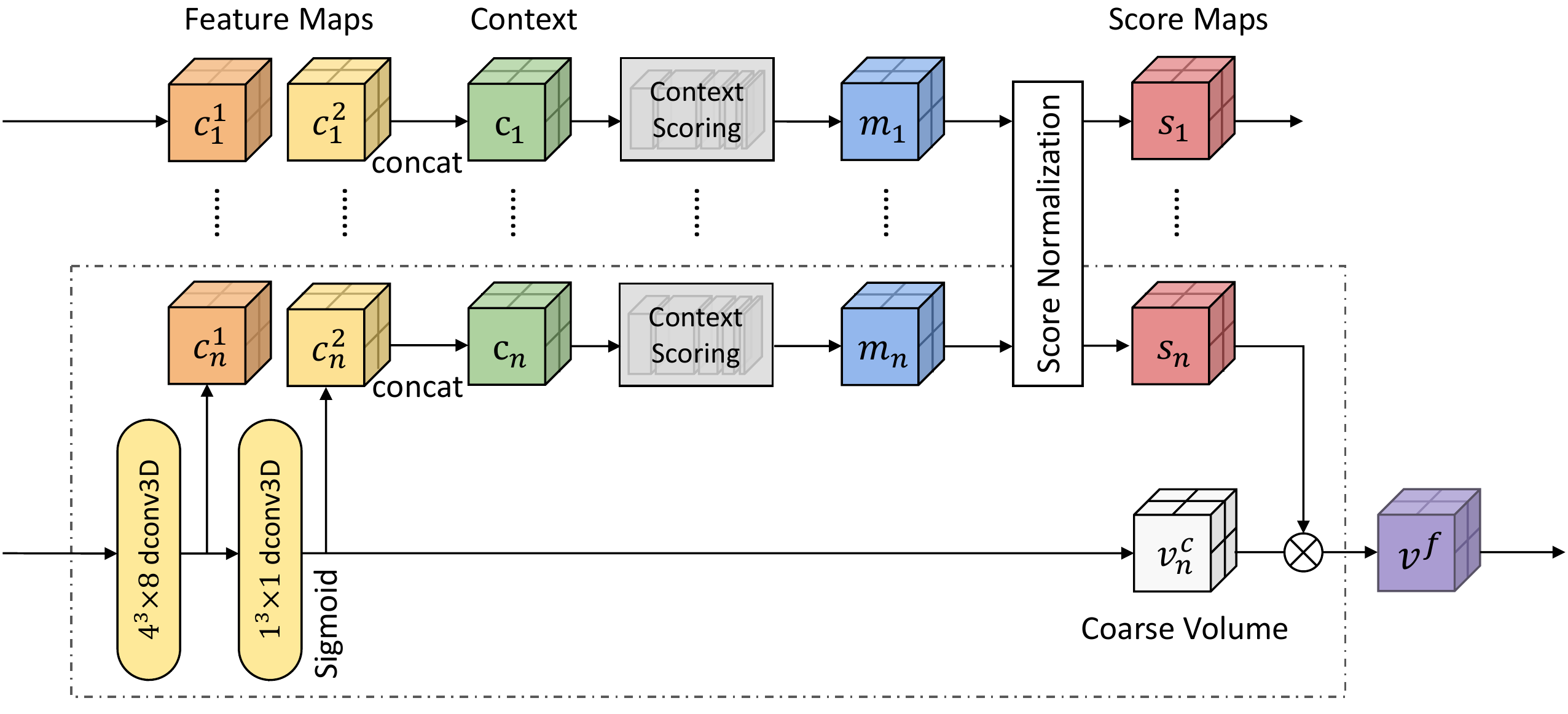}
  }
  \caption{An overview of the context-aware fusion module. It aims to select high-quality reconstructions for each part to construct the final results. The objects in the bounding box describe the procedure score calculation for a coarse volume $v^c_n$. The other scores are calculated according to the same procedure. Note that the weights of the context scoring network are shared among different views.}
  \label{fig:context-aware-fusion}
  \vspace{-2 mm}
\end{figure}

From different viewpoints, we can see different visible parts of an object.
The reconstruction qualities of visible parts are much higher than those of invisible parts.
Inspired by this observation, we propose a context-aware fusion module to adaptively select high-quality reconstruction for each part (e.g., table legs) from different coarse 3D volumes.
The selected reconstructions are fused to generate a 3D volume of the whole object (Figure \ref{fig:context-aware-fusion-visualization}).

As shown in Figure \ref{fig:context-aware-fusion}, given coarse 3D volumes and the corresponding context, the context-aware fusion module generates a score map for each coarse volume and then fuses them into one volume by the weighted summation of all coarse volumes according to their score maps.
The spatial information of voxels is preserved in the context-aware fusion module, and thus Pix2Vox can utilize multi-view information to recover the structure of an object better.

Specifically, the context-aware fusion module generates the context $c_r$ of the $r$-th coarse volume $v_r^c$ by concatenating the output of the last two layers in the decoder.
Then, the context scoring network generates a score $m_r$ for the context of the $r$-th coarse voxel.
The context scoring network is composed of five sets of 3D convolutional layers, each of which has a kernel size of $3^3$ and padding of $1$, followed by a batch normalization and a leaky ReLU activation.
The numbers of output channels of convolutional layers are $9$, $16$, $8$, $4$, and $1$, respectively.
The learned score $m_r$ for context $c_r$ are normalized across all learnt scores.
We choose softmax as the normalization function. 
Therefore, the score $s_r^{(i, j, k)}$ at position $(i, j, k)$ for the $r$-th voxel can be calculated as 

\begin{equation}
  s_r^{(i, j, k)} = \frac{\exp\left(m_r^{(i, j, k)}\right)}{\sum_{p=1}^n \exp\left(m_p^{(i, j, k)}\right)}  
\end{equation}
where $n$ represents the number of views.
Finally, the fused voxel $v^f$ is produced by summing up the product of coarse voxels and the corresponding scores altogether.

\begin{equation}
  v^f = \sum_{r=1}^n s_r v_r^c
\end{equation}

\vspace{-3 mm}
\subsubsection{Refiner}
\vspace{-1 mm}

The refiner can be seen as a residual network, which aims to correct wrongly recovered parts of a 3D volume.
It follows the idea of a 3D encoder-decoder with the U-net connections \cite{DBLP:conf/miccai/RonnebergerFB15}. 
With the help of the U-net connections between the encoder and decoder, the local structure in the fused volume can be preserved.
Specifically, the encoder has three 3D convolutional layers, each of which has a bank of $4^3$ filters with padding of $2$, followed by a batch normalization layer, a leaky ReLU activation and a max pooling layer with a kernel size of $2^3$.
The numbers of output channels of convolutional layers are $32$, $64$, and $128$, respectively.
The encoder is finally followed by two fully connected layers with dimensions of $2048$ and $8192$.
The decoder consists of three transposed convolutional layers, each of which has a bank of $4^3$ filters with padding of $2$ and stride of $1$.

Except for the last transposed convolutional layer that is followed by a sigmoid function, other layers are followed by a batch normalization layer and a ReLU activation.

\begin{table*}[!t]
  \caption{Single-view reconstruction on ShapeNet compared using Intersection-over-Union (IoU). The best number for each category is highlighted in bold. Note that DRC \cite{DBLP:conf/cvpr/TulsianiZEM17} is trained/tested per category and PSGN \cite{DBLP:conf/cvpr/FanSG17} takes object masks as an additional input. Besides, PSGN uses 220k 3D CAD models while the remaining methods use only 44k 3D CAD models during training.}
  \vspace{-2 mm}
  \centering
  \begin{tabularx}{\linewidth}{XYYYYYY}
    \toprule
    Category     & 3D-R2N2 \cite{DBLP:conf/eccv/ChoyXGCS16}
                 & OGN \cite{DBLP:conf/iccv/TatarchenkoDB17}
                 & DRC \cite{DBLP:conf/cvpr/TulsianiZEM17}
                 & PSGN \cite{DBLP:conf/cvpr/FanSG17}
                 & Pix2Vox-F & Pix2Vox-A \\
    \midrule
    airplane     & 0.513      & 0.587      & 0.571
                 & 0.601      & 0.600      & \bf{0.684} \\
    bench        & 0.421      & 0.481      & 0.453
                 & 0.550      & 0.538      & \bf{0.616} \\
    cabinet      & 0.716      & 0.729      & 0.635
                 & 0.771      & 0.765      & \bf{0.792} \\
    car          & 0.798      & 0.828      & 0.755
                 & 0.831      & 0.837      & \bf{0.854} \\
    chair        & 0.466      & 0.483      & 0.469
                 & 0.544      & 0.535      & \bf{0.567} \\
    display      & 0.468      & 0.502      & 0.419
                 & \bf{0.552} & 0.511      & 0.537 \\
    lamp         & 0.381      & 0.398      & 0.415
                 & \bf{0.462} & 0.435      & 0.443 \\
    speaker      & 0.662      & 0.637      & 0.609
                 & \bf{0.737} & 0.707      & 0.714 \\
    rifle        & 0.544      & 0.593      & 0.608
                 & 0.604      & 0.598      & \bf{0.615} \\
    sofa         & 0.628      & 0.646      & 0.606
                 & 0.708      & 0.687      & \bf{0.709} \\
    table        & 0.513      & 0.536      & 0.424
                 & \bf{0.606} & 0.587      & 0.601 \\
    telephone    & 0.661      & 0.702      & 0.413
                 & 0.749      & 0.770      & \bf{0.776} \\
    watercraft   & 0.513      & \bf{0.632} & 0.556
                 & 0.611      & 0.582      & 0.594  \\
    \midrule
    Overall      & 0.560      & 0.596      & 0.545
                 & 0.640      & 0.634      & \bf{0.661} \\
    \bottomrule
  \end{tabularx}
  \label{tab:shapenet-reconstruction}
\end{table*}

\begin{table*}[!t]
  \caption{Multi-view reconstruction on ShapeNet compared using Intersection-over-Union (IoU). The best results for different numbers of views are highlighted in bold. The marker $^\dag$ indicates that the context-aware fusion is replaced with the average fusion.}
  \vspace{-2 mm}
  \centering
  \begin{tabularx}{\linewidth}{lYYYYYYYYY}
    \toprule
    Methods   & 1 view         & 2 views        & 3 views
              & 4 views        & 5 views        & 8 views
              & 12 views       & 16 views       & 20 views \\
    \midrule
    3D-R2N2   \cite{DBLP:conf/eccv/ChoyXGCS16}
              & 0.560          & 0.603          & 0.617
              & 0.625          & 0.634          & 0.635
              & 0.636          & 0.636          & 0.636 \\
    Pix2Vox-F \hspace{.6mm}$^\dag$ 
              & 0.634          & 0.653          & 0.661
              & 0.666          & 0.668          & 0.672
              & 0.674          & 0.675          & 0.676 \\
    Pix2Vox-F 
              & 0.634          & 0.660          & 0.668
              & 0.673          & 0.676          & 0.680
              & 0.682          & 0.684          & 0.684 \\
    Pix2Vox-A $^\dag$
              & \textbf{0.661} & 0.678          & 0.684
              & 0.687          & 0.689          & 0.692
              & 0.694          & 0.695          & 0.695 \\
    Pix2Vox-A & \textbf{0.661} & \textbf{0.686} & \textbf{0.693}
              & \textbf{0.697} & \textbf{0.699} & \textbf{0.702}
              & \textbf{0.704} & \textbf{0.705} & \textbf{0.706} \\
    \bottomrule
  \end{tabularx}
  \label{tab:shapenet-multi-view-reconstruction}
\end{table*}

\vspace{-3 mm}
\subsubsection{Loss Function}
\vspace{-1 mm}

The loss function of the network is defined as the mean value of the voxel-wise binary cross entropies between the reconstructed object and the ground truth.
More formally, it can be defined as

\begin{equation}
  \ell = \frac{1}{N} \sum_{i=1}^N \left[ gt_i \log(p_i) + (1 - gt_i) \log(1 - p_i) \right]
  \label{eq:bce-loss}
\end{equation}
where $N$ denotes the number of voxels in the ground truth. $p_i$ and $gt_i$ represent the predicted occupancy and the corresponding ground truth.
The smaller the $\ell$ value is, the closer the prediction is to the ground truth.

\section{Experiments}

\begin{figure*}
  \centering
  \resizebox{\linewidth}{!} {
    \includegraphics{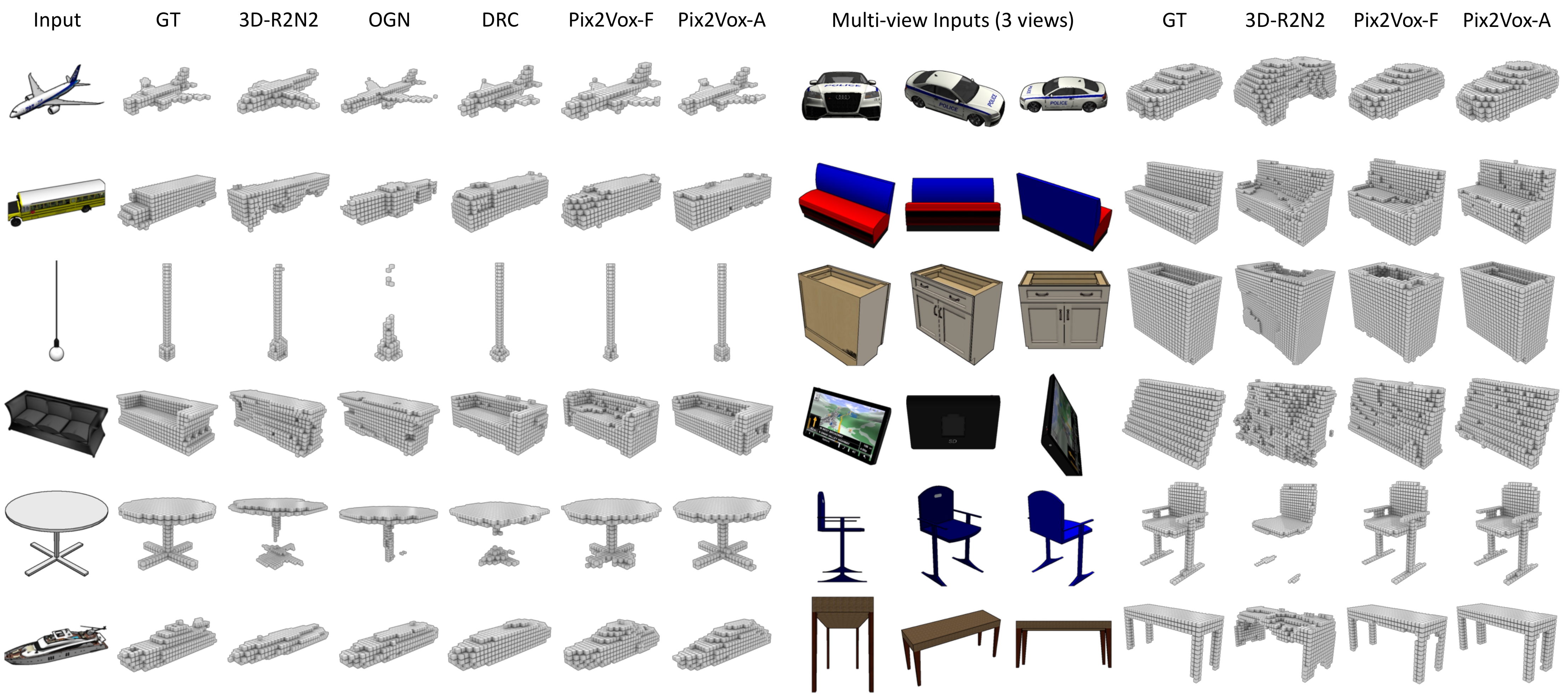}
  }
  \caption{Single-view (left) and multi-view (right) reconstructions on the ShapeNet testing set. GT represents the ground truth of the 3D object. Note that DRC \cite{DBLP:conf/cvpr/TulsianiZEM17} is trained/tested per category.}
  \label{fig:shapenet-reconstruction}
  \vspace{-2 mm}
\end{figure*}

\subsection{Datasets and Metrics}

\noindent \textbf{Datasets}
We evaluate the proposed Pix2Vox-F and Pix2Vox-A on both synthetic images of objects from the ShapeNet \cite{DBLP:conf/cvpr/WuSKYZTX15} dataset and real images from the Pix3D \cite{DBLP:conf/cvpr/Sun0ZZZXTF18} dataset.
More specifically, we use a subset of ShapeNet consisting of 13 major categories and 43,783 3D models following the settings of \cite{DBLP:conf/eccv/ChoyXGCS16}.
As for Pix3D, we use the 2,894 untruncated and unoccluded chair images following the settings of \cite{DBLP:conf/cvpr/Sun0ZZZXTF18}.

\noindent \textbf{Evaluation Metrics}
To evaluate the quality of the output from the proposed methods, we binarize the probabilities at a fixed threshold of 0.3 and use intersection over union (IoU) as the similarity measure.
More formally,

\begin{equation}
  {\rm IoU} = \frac{\sum_{i, j, k} {\rm I}(p_{(i, j, k)} > t) {\rm I}(gt_{(i, j, k)})}{\sum_{i, j, k} {\rm I}\left[ {\rm I}(p_{(i, j, k)} > t) + {\rm I}(gt_{(i, j, k)}) \right]}
\end{equation}
where $p_{(i, j, k)}$ and $gt_{(i, j, k)}$ represent the predicted occupancy probability and the ground truth at $(i, j, k)$, respectively.
${\rm I}(\cdot)$ is an indicator function and $t$ denotes a voxelization threshold.
Higher IoU values indicate better reconstruction results.

\subsection{Implementation Details}

We use $224 \times 224$ RGB images as input to train the proposed methods with a shape batch size of $64$.
The output voxelized reconstruction is $32^3$ in size.
We implement our network in PyTorch \cite{DBLP:conf/nips/AdamSSGEZZALA17} and train both Pix2Vox-F and Pix2Vox-A using an Adam optimizer \cite{DBLP:conf/iclr/KingmaB15} with a $\beta_1$ of $0.9$ and a $\beta_2$ of $0.999$.
The initial learning rate is set to $0.001$ and decayed by 2 after 150 epochs.
First, we train both networks except the context-aware fusion feeding with a single-view image for 250 epochs.
Then, we train the whole network jointly feeding with random numbers of input images for 100 epochs.

\subsection{Reconstruction of Synthetic Images}
\label{sec:shapenet-reconstruction}

To evaluate the performance of the proposed methods in handling synthetic images, we compare our methods against several state-of-the-art methods on the ShapeNet testing set.
To make a fair comparison, all methods are compared with the same input images for all experiments except PSGN \cite{DBLP:conf/cvpr/FanSG17}.
Although PSGN uses much more data during training, Pix2Vox-A still performs better in recovering the 3D shape of an object.
Table \ref{tab:shapenet-reconstruction} shows the performance of single-view reconstruction, while Table \ref{tab:shapenet-multi-view-reconstruction} shows the mean IoU scores of multi-view reconstruction with different numbers of views.

The single-view reconstruction results of Pix2Vox-F and Pix2Vox-A significantly outperform other methods (Table \ref{tab:shapenet-reconstruction}).
Pix2Vox-A increases IoU over 3D-R2N2 by 18\%. 
In multi-view reconstruction, Pix2Vox-A consistently outperforms 3D-R2N2 in all numbers of views (Table \ref{tab:shapenet-multi-view-reconstruction}). 
The IoU of Pix2Vox-A is 13\% higher than that of 3D-R2N2.

Figure \ref{fig:shapenet-reconstruction} shows several reconstruction examples from the ShapeNet testing set.
Both Pix2Vox-F and Pix2Vox-A are able to recover the thin parts of objects, such as lamps and table legs.
Compare with Pix2Vox-F, we also observe that higher dimensional feature maps in Pix2Vox-A do contribute to 3D reconstruction.
Moreover, in multi-view reconstruction, both Pix2Vox-A and Pix2Vox-F produce better results than 3D-R2N2.

\subsection{Reconstruction of Real-world Images}
\label{sec:pix3d-reconstruction}

\begin{figure}[!t]
  \centering
  \resizebox{\linewidth}{!} {
    \includegraphics{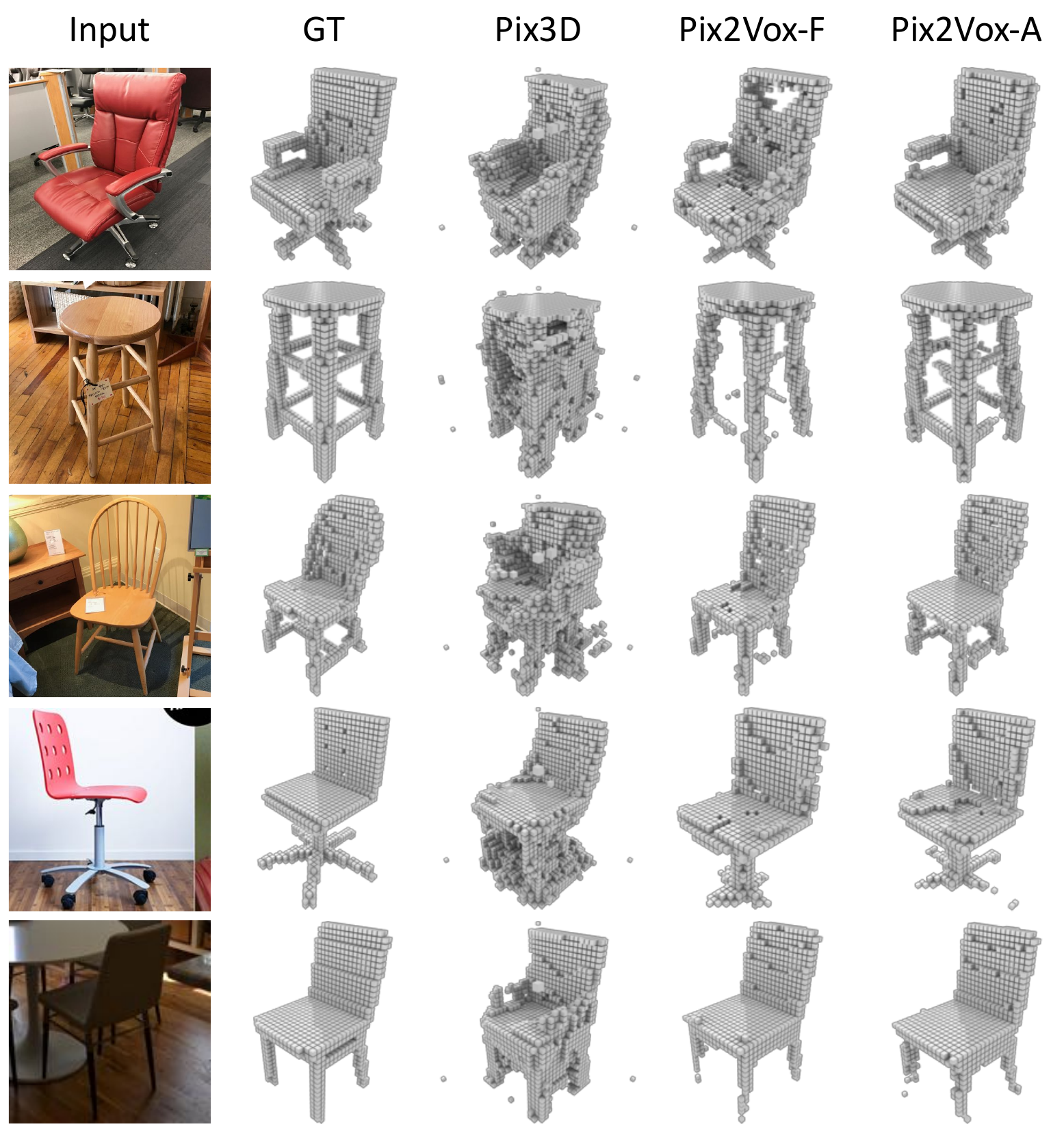}
  }
  \caption{Reconstruction on the Pix3D testing set from single-view images. GT represents the ground truth of the 3D object.}
  \label{fig:pix3d-reconstruction}
  \vspace{-2 mm}
\end{figure}

\begin{table}
  \caption{Single-view reconstruction on Pix3D compared using Intersection-over-Union (IoU). The best number is highlighted in bold.}
  \vspace{-2 mm}
  \centering
  \begin{tabularx}{\linewidth}{XY}
    \toprule
    ~Method                                              & IoU \\
    \midrule
    ~3D-R2N2 \cite{DBLP:conf/eccv/ChoyXGCS16}            & 0.136 \\
    ~DRC \cite{DBLP:conf/cvpr/TulsianiZEM17}             & 0.265 \\
    ~Pix3D (w/o Pose) \cite{DBLP:conf/cvpr/Sun0ZZZXTF18} & 0.267 \\
    ~Pix3D (w/ Pose) \cite{DBLP:conf/cvpr/Sun0ZZZXTF18}  & 0.282 \\
    ~Pix2Vox-F                                           & 0.271 \\
    ~Pix2Vox-A                                           & \bf{0.288}\\
    \bottomrule
  \end{tabularx}
  \label{tab:pix3d-reconstruction}
  \vspace{-2 mm}
\end{table}

To evaluate the performance on of the proposed methods on real-world images, we test our methods for single-view reconstruction on the Pix3D dataset.

We use the pipeline of RenderForCNN \cite{DBLP:conf/iccv/SuQLG15} to generate 60 images for each 3D CAD model in the ShapeNet dataset.
We perform quantitative evaluation of the resulting models on real-world RGB images using the Pix3D dataset.
Besides, we augment our training data by random color and light jittering.
First, the images are cropped according to the bounding box of the objects within the image.
Then, these cropped images are rescaled as required by each reconstruction network.

The mean IoU of the Pix3D dataset is reported in Table \ref{tab:pix3d-reconstruction}.
The experimental results indicate Pix2Vox-A outperform the competing approaches on the Pix3D testing set without estimating the pose of an object.
The qualitative analysis is given in Figure \ref{fig:pix3d-reconstruction}, which indicate that the proposed methods are more effective in handling real-world scenarios.

\subsection{Reconstruction of Unseen Objects}

\begin{figure}
  \centering
  \resizebox{\linewidth}{!} {
    \includegraphics{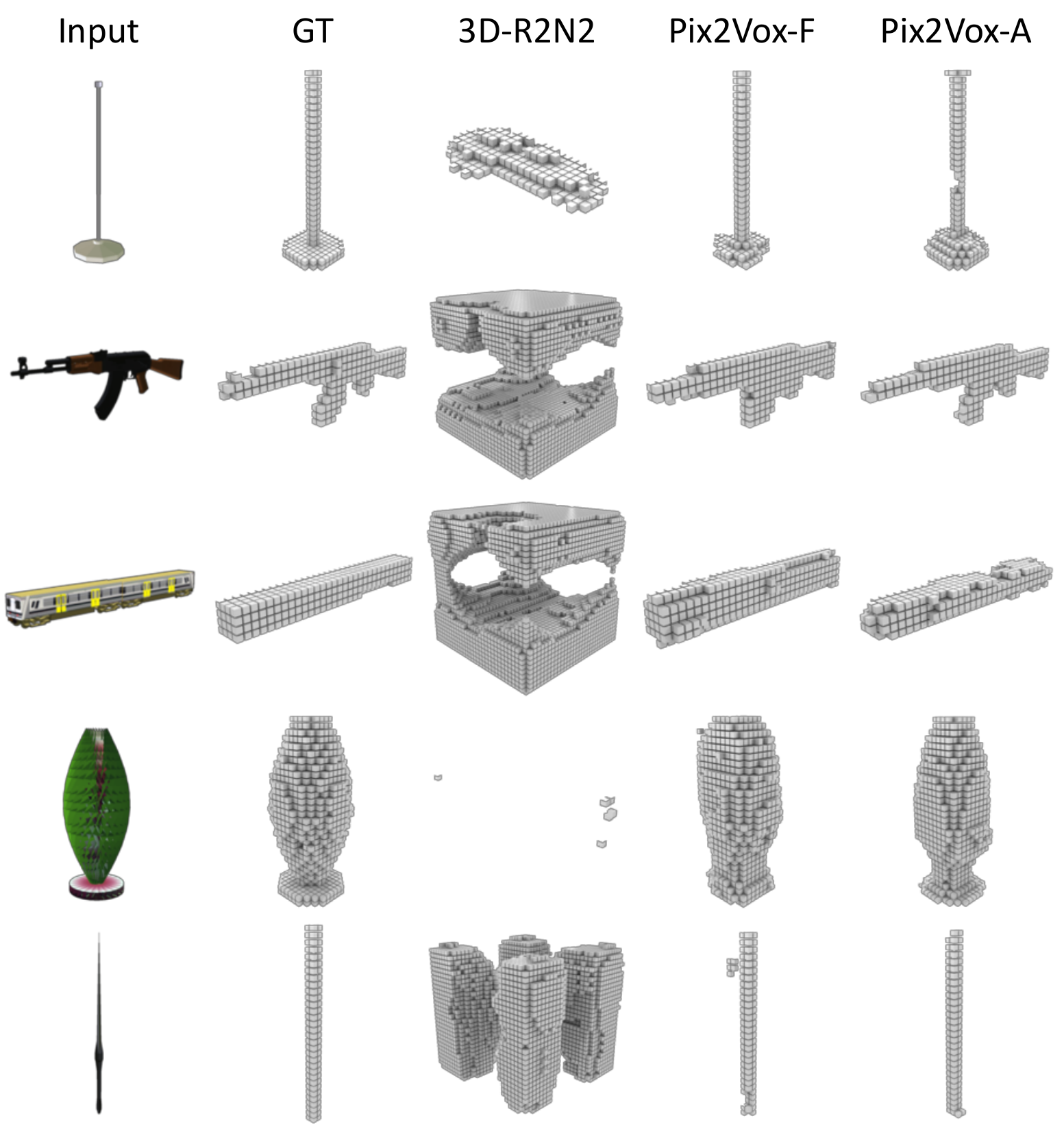}
  }
  \caption{Reconstruction on unseen objects of ShapeNet from 5-view images. GT represents the ground truth of the 3D object.}
  \label{fig:shapenet-unseen-reconstruction}
  \vspace{-2 mm}
\end{figure}

In order to test how well our methods can generalize to unseen objects, we conduct additional experiments on ShapeNetCore \cite{DBLP:conf/cvpr/WuSKYZTX15}.
We use Mitsuba\footnote{https://www.mitsuba-renderer.org} to render objects in the remaining 44 categories of ShapeNetCore from 24 random views along with voxel representations.
All pretrained models have never ``seen'' either the objects in these categories or the labels of objects before.
More specifically, all models are trained on the 13 major categories of ShapeNet renderings provided by \cite{DBLP:conf/eccv/ChoyXGCS16} and tested on the remaining 44 categories of ShapeNetCore with the same input images.
The reconstruction results of 3D-R2N2 are obtained with the released pretrained model.

Several reconstruction results are presented in Figure \ref{fig:shapenet-unseen-reconstruction}.
The reconstruction IoU of 3D-R2N2 on unseen objects is $0.120$, while Pix2Vox-F and Pix2Vox-A are $0.209$ and $0.227$, respectively.
Experimental results demonstrate that 3D-R2N2 can hardly recover the shape of unseen objects.
In contrast, Pix2Vox-F and Pix2Vox-A show satisfactory generalization abilities to unseen objects.

\subsection{Ablation Study}

In this section, we validate the context-aware fusion and the refiner by ablation studies.

\noindent \textbf{Context-aware fusion}
To quantitatively evaluate the context-aware fusion, we replace the context-aware fusion in Pix2Vox-A with the average fusion, where the fused voxel $v^f$ can be calculated as

\begin{equation}
  v^f_{(i, j, k)} = \frac{1}{n} \sum_{r=1}^n v^r_{(i, j, k)}
\end{equation}
Table \ref{tab:shapenet-multi-view-reconstruction} shows that the context-aware fusion performs better than the average fusion in selecting the high-quality reconstructions for each part from different coarse volumes.

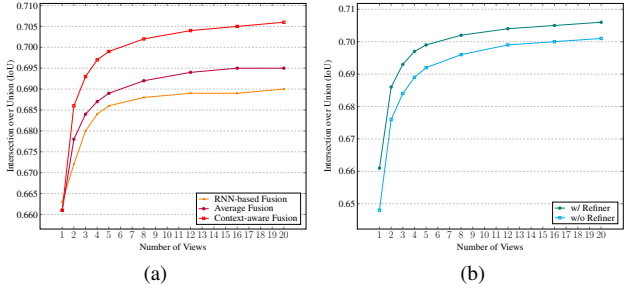
\begin{figure}[!t]
  \centering
  \subfloat[]{
    \resizebox{.485\linewidth}{!} {
      \begin{tikzpicture}[every plot/.append style={ultra thick}]
        \begin{axis} [
          width = \textwidth,
          xlabel = {Number of Views},
          xlabel style = {yshift=-0.2cm, font=\fontsize{16}{16}\selectfont},
          ylabel = {Intersection over Union (IoU)},
          ylabel style = {yshift=0.5cm, font=\fontsize{16}{16}\selectfont},
          log basis x = {2},
          xtick = {1, 2, 3, 4, 5, 6, 7, 8, 9, 10, 11, 12, 13, 14, 15, 16, 17, 18, 19, 20},
          xticklabel style = {font=\fontsize{16}{16}\selectfont},
          ytick = {0.650, 0.655, 0.66, 0.665, 0.670, 0.675, 0.680, 0.685, 0.690, 0.695, 0.700, 0.705, 0.710},
          yticklabels = {0.650, 0.655, 0.660, 0.665, 0.670, 0.675, 0.680, 0.685, 0.690, 0.695, 0.700, 0.705, 0.710},
          yticklabel style = {font=\fontsize{16}{16}\selectfont},
          legend style = {at={(0.98, 0.16)}, column sep=0.25cm, font=\fontsize{16}{16}\selectfont},
          legend cell align = left,
          ymajorgrids = true,
          grid style = dashed,
        ]

        \addplot[
          color=orange,
          mark=x,
        ]
        coordinates {
          (1, 0.663)(2, 0.672)(3, 0.680)(4, 0.684)(5, 0.686)
          (8, 0.688)(12, 0.689)(16, 0.689)(20, 0.690)
        };
        \addlegendentry{RNN-based Fusion};

        \addplot[
          color=purple,
          mark=*,
        ]
        coordinates {
          (1, 0.661)(2, 0.678)(3, 0.684)(4, 0.687)(5, 0.689)
          (8, 0.692)(12, 0.694)(16, 0.695)(20, 0.695)
        };
        \addlegendentry{Average Fusion};

        \addplot[
          color=red,
          mark=square,
        ]
        coordinates {
          (1, 0.661)(2, 0.686)(3, 0.693)(4, 0.697)(5, 0.699)
          (8, 0.702)(12, 0.704)(16, 0.705)(20, 0.706)
        };
        \addlegendentry{Context-aware Fusion};
        \end{axis}
        \label{fig:ablation-context-aware-fusion}
      \end{tikzpicture}
    }
  }
  \subfloat[]{
    \resizebox{.485\linewidth}{!} {
      \begin{tikzpicture}[every plot/.append style={ultra thick}]
        \begin{axis} [
          width = \textwidth,
          xlabel = {Number of Views},
          xlabel style = {yshift=-0.2cm, font=\fontsize{16}{16}\selectfont},
          ylabel = {Intersection over Union (IoU)},
          ylabel style = {yshift=0.5cm, font=\fontsize{16}{16}\selectfont},
          log basis x = {2},
          xtick = {1, 2, 3, 4, 5, 6, 7, 8, 9, 10, 11, 12, 13, 14, 15, 16, 17, 18, 19, 20},
          xticklabel style = {font=\fontsize{16}{16}\selectfont},
          ytick = {0.63, 0.64, 0.65, 0.66, 0.67, 0.68, 0.69, 0.70, 0.71},
          yticklabels = {0.63, 0.64, 0.65, 0.66, 0.67, 0.68, 0.69, 0.70, 0.71},
          yticklabel style = {font=\fontsize{16}{16}\selectfont},
          legend style = {at={(0.98, 0.12)}, column sep=0.25cm, font=\fontsize{16}{16}\selectfont},
          legend cell align = left,
          ymajorgrids = true,
          grid style = dashed,
        ]

        \addplot[
          color=teal,
          mark=*,
        ]
        coordinates {
          (1, 0.661)(2, 0.686)(3, 0.693)(4, 0.697)(5, 0.699)
          (8, 0.702)(12, 0.704)(16, 0.705)(20, 0.706)
        };
        \addlegendentry{w/ Refiner};

        \addplot[
          color=cyan,
          mark=square,
        ]
        coordinates {
          (1, 0.648)(2, 0.676)(3, 0.684)(4, 0.689)(5, 0.692)
          (8, 0.696)(12, 0.699)(16, 0.700)(20, 0.701)
        };
        \addlegendentry{w/o Refiner};
        \end{axis}
        \label{fig:ablation-refiner}
      \end{tikzpicture}
    }
  }
  \caption{The IoU on ShapeNet testing set. (a) Effects of the context aware fusion and the number of views on the evaluation IoU. (b) Effects of the refiner network and the number of views on the evaluation IoU. }
\end{figure}

To make a further comparison with RNN-based fusion, we remove the context-aware fusion and add an 3D convolutional LSTM \cite{DBLP:conf/eccv/ChoyXGCS16} after the encoder.
To fit the input of the 3D convolutional LSTM, we add an additional fully-connected layer with a dimension of $1024$ before it.
As shown in Figure \ref{fig:ablation-context-aware-fusion}, both the average fusion and context-aware fusion consistently outperform the RNN-based fusion in all numbers of views. 

\noindent \textbf{Refiner}
Pix2Vox-A uses a refiner to further refine the fused 3D volume.
For single-view reconstruction on ShapeNet, the IoU of Pix2Vox-A is $0.661$. 
In contrast, the IoU of Pix2Vox-A without the refiner decreases to $0.636$.
Removing refiner causes considerable degeneration for the reconstruction accuracy.
As shown in Figure \ref{fig:ablation-refiner}, as the number of views increases, the effect of the refiner becomes weaker.

The ablation studies indicate that both the context-aware fusion and the refiner play important roles in our framework for the performance improvements against previous state-of-the-art methods.

\subsection{Space and Time Complexity}
\begin{table}
  \caption{Memory usage and running time on ShapeNet dataset. Note that backward time is measured in single-view reconstruction with a batch size of 1.}
  \vspace{-2 mm}
  \centering
  \resizebox{\linewidth}{!} {
    \begin{tabular}{lccccc}
    \toprule
    Methods               & 3D-R2N2   & OGN
                          & Pix2Vox-F & Pix2Vox-A \\
    \midrule
    \#Parameters (M)      & 35.97     & 12.46
                          & 7.41      & 114.24 \\
    Memory (MB)           & 1407      & 793
                          & 673       & 2729 \\
    \midrule
    Training (hours)      & 169       & 192
                          & 12        & 25 \\
    Backward (ms)         & 312.50    & 312.25
                          & 12.93     & 72.01 \\
    \midrule
    Forward, 1-view (ms)  & 73.35     & 37.90
                          & 9.25      & 9.90 \\
    Forward, 2-views (ms) & 108.11    & N/A
                          & 12.05     & 13.69 \\
    Forward, 4-views (ms) & 112.36    & N/A
                          & 23.26     & 26.31 \\
    Forward, 8-views (ms) & 117.64    & N/A
                          & 52.63     & 55.56 \\
    \bottomrule
    \end{tabular}
  }
  \label{tab:performance-comparison}
  \vspace{-2 mm}
\end{table}

Table \ref{tab:performance-comparison} and Figure \ref{fig:performance-comparison} show the numbers of parameters of different methods.
There is an $80\%$ reduction in parameters in Pix2Vox-F compared to 3D-R2N2.

The running times are obtained on the same PC with an NVIDIA GTX 1080 Ti GPU.
For more precise timing, we exclude the reading and writing time when evaluating the forward and backward inference time.
Both Pix2Vox-F and Pix2Vox-A are about $8$ times faster in forward inference than 3D-R2N2 in single-view reconstruction.
In backward inference, Pix2Vox-F and Pix2Vox-A are about $24$ and $4$ times faster than 3D-R2N2, respectively.

\subsection{Discussion}

To give a detailed analysis of the context-aware fusion module, we visualized the score maps of three coarse volumes when reconstructing the 3D shape of a table from 3-view images, as shown in Figure \ref{fig:context-aware-fusion-visualization}.
The reconstruction quality of the table tops on the right is clearly of low quality, and the score of the corresponding part is lower than those in the other two coarse volumes.
The fused 3D volume is obtained by combining the selected high-quality reconstruction parts, where bad reconstructions can be eliminated effectively by our scoring scheme.

Pix2Vox recovers the 3D shape of an object without knowing camera parameters.
To further demonstrate the superior ability of the context-aware fusion in multi-view stereo (MVS) systems \cite{DBLP:conf/cvpr/SeitzCDSS06}, we replace the RNN with the context-aware fusion in LSM \cite{DBLP:conf/nips/KarHM17}.
Specifically, we remove the recurrent fusion and add the context-aware fusion to combine the 3D volume reconstruction of each view.
Experimental results show that the IoU is increased by about 2\% on the ShapeNet testing set, which indicate that the context-aware fusion also helps MVS systems to obtain better reconstruction results.

Although our methods outperform state-of-the-arts, the reconstruction results of our methods are still with a low resolution.
We can further improve the reconstruction resolutions in the future work by introducing GANs \cite{DBLP:conf/nips/GoodfellowPMXWOCB14}.

\section{Conclusion and Future Works}

In this paper, we propose a unified framework for both single-view and multi-view 3D reconstruction, named Pix2Vox.
Compared with existing methods that fuse deep features generated by a shared encoder, the proposed method fuses multiple coarse volumes produced by a decoder and preserves multi-view spatial constraints better.
Quantitative and qualitative evaluation for both single-view and multi-view reconstruction on the ShapeNet and Pix3D benchmarks indicate that the proposed methods outperform state-of-the-arts by a large margin.
Pix2Vox is computationally efficient, which is 24 times faster than 3D-R2N2 in terms of backward inference time.
In future work, we will work on improving the resolution of the reconstructed 3D objects.
In addition, we also plan to extend Pix2Vox to reconstruct 3D objects from RGB-D images.

\noindent \textbf{Acknowledgements}
This work was supported by the National Natural Science Foundation of China under Project No. 61772158, 61702136, 61872112 and U1711265.
We gratefully acknowledge Prof. Junbao Li and Huanyu Liu for providing additional GPU hours for this research.
We would also like to thank Prof. Wangmeng Zuo, Jiapeng Tang, and anonymous reviewers for their valuable feedbacks and help during this research.

{\small
\bibliographystyle{ieee}
\bibliography{references}
}

\end{document}